\documentclass[10pt,twocolumn,letterpaper]{article}

\usepackage{cvpr}
\usepackage{times}
\usepackage{epsfig}
\usepackage{graphicx}
\usepackage{amsmath}
\usepackage{amssymb}
\usepackage{dsfont}
\usepackage{subcaption}

\usepackage{enumitem}
\usepackage{booktabs}
\usepackage{multirow}
\usepackage{pdfpages}

\usepackage[pagebackref=true,breaklinks=true,letterpaper=true,colorlinks,bookmarks=false]{hyperref}

\cvprfinalcopy 

\def\cvprPaperID{5862} 

\graphicspath{{figures_compressed/}}

\newcommand{\boldparagraph}[1]{\vspace{0.2em}\noindent{\bf #1} }

\newcommand{\sfc}[1]{{#1}}  

\ifcvprfinal\fi

\begin{document}

\title{RoutedFusion: Learning Real-time Depth Map Fusion}

\author{Silvan Weder$^1$
			\qquad
			Johannes L.~Sch\"{o}nberger$^2$
			\qquad
			Marc Pollefeys$^{1,2}$
			\qquad
			Martin R.~Oswald$^1$\\
			$^1$Department of Computer Science, ETH Zurich\\
			$^2$Microsoft}


\maketitle

\begin{abstract}
The efficient fusion of depth maps is a key part of most state-of-the-art 3D reconstruction methods. Besides requiring high accuracy, these depth fusion methods need to be scalable and real-time capable.
To this end, we present a novel real-time capable machine learning-based method for depth map fusion.
Similar to the seminal depth map fusion approach by Curless and Levoy,
we only update a local group of voxels to ensure real-time capability.
Instead of a simple linear fusion of depth information, we propose a neural network that predicts non-linear updates to better account for typical fusion errors.
Our network is composed of a 2D depth routing network and a 3D depth fusion network which efficiently handle sensor-specific noise and outliers.
This is especially useful for surface edges and thin objects for which the original approach suffers from thickening artifacts.
Our method outperforms the traditional fusion approach and related learned approaches on both synthetic and real data. 
We demonstrate the performance of our method in reconstructing fine geometric details from noise and outlier contaminated data on various scenes.
\end{abstract}

\section{Introduction}
\label{sec:introduction}

Multi-view 3D reconstruction has been a central research topic in computer vision for many decades.
Fusing depth maps from multiple camera viewpoints is an essential processing step in the majority of recent 3D reconstruction pipelines~\cite{Zach-et-al-ICCV-2007,Zach-3DPVT-2008,Kolev-et-al-IJCV-2009,Blaha-et-al-CVPR-2016,Savinov-et-al-CVPR-2015,Savinov-et-al-CVPR-2016,Dai-et-al-CVPR-2018,Dai-Niessner-ECCV-2018}, especially for real-time applications~\cite{Izadi-et-al-SIGGRAPH-2011,Niessner-et-al-SIGGRAPH-2013,Whelan-et-al-IJRR-2016,Dai-et-et-al-SIGGRAPH-2017}.
%
We revisit the problem of 3D reconstruction via depth map fusion from a machine learning perspective.
The major difficulty of this task is to deal with various amounts of noise, outliers, and missing data.
The classical approach~\cite{Curless-Levoy-SIGGRAPH-1996,Izadi-et-al-SIGGRAPH-2011} to fusing noisy depth maps is to average truncated signed distance functions (TSDF).
This approach has many advantages:
\mbox{\textbf{1+)}} The updates are local (truncated) and can be done in constant time for a fixed number of depth values.
The high memory usage of voxel grids can be easily reduced with voxel hashing~\cite{Niessner-et-al-SIGGRAPH-2013} or octrees~\cite{Steinbruecker-et-al-ICCV-2013}.
\mbox{\textbf{2+)}} Online updates are simple to implement and noisy measurements are fused into a single surface with very few operations. 
\mbox{\textbf{3+)}} Due to local independent updates, the approach is computationally cheap and highly parallelizable.

%
\begin{figure}[t!]
  \small
  \centering
  \setlength{\tabcolsep}{0.5mm}
  \newcommand{\sz}{0.48}
  \begin{tabular}{cc}
  	\includegraphics[width=\sz\columnwidth,trim={98, 62, 98, 62},clip]{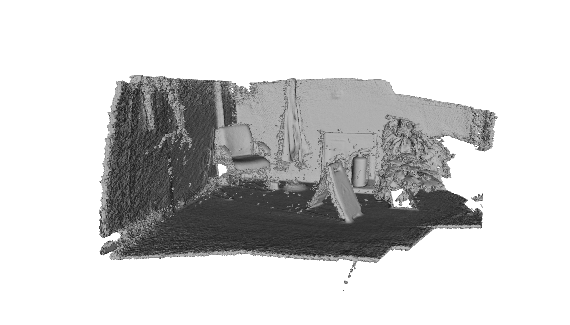} &
  	\includegraphics[width=\sz\columnwidth,trim={98, 62, 98, 62},clip]{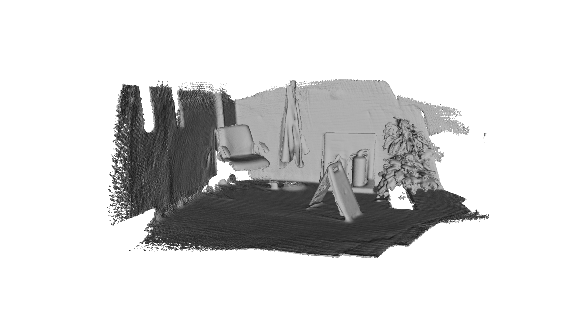} \\
  	\includegraphics[width=\sz\columnwidth,trim={70, 0, 112, 0},clip]{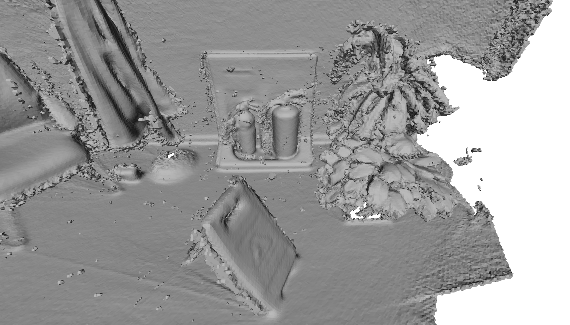} &
  	\includegraphics[width=\sz\columnwidth,trim={70, 0, 112, 0},clip]{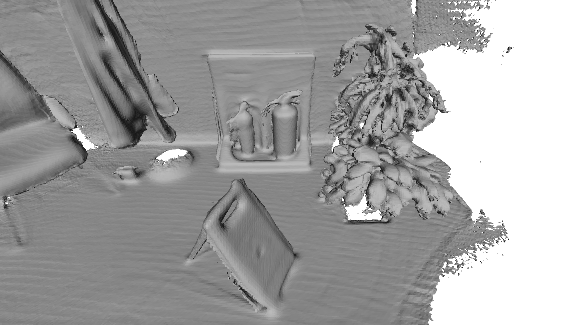} \\
    \sfc{Standard TSDF Fusion~\cite{Curless-Levoy-SIGGRAPH-1996}} & \sfc{Ours} \\[-6pt]
  \end{tabular}
  \caption{\textbf{Standard TSDF fusion vs.~our learned depth map fusion approach} (on Kinect data~\cite{7scene}).
	Due to a more informed decision process, our approach better handles noise and fine geometric details.}
  \label{fig:teaser}
\end{figure}

However, the approach also has a number of shortcomings:
\mbox{\textbf{1-)}} The average is only the optimal estimate for zero-mean Gaussian noise, but the real error distribution is typically non-Gaussian, non-centered and depth-dependent.
\mbox{\textbf{2-)}} The updates are linear and a minimal thickness assumption of surfaces has to be made according to the expected noise level. Therefore, thickening artifacts become apparent along surface edges and for thin object structures.
\mbox{\textbf{3-)}} This issue is even more severe when depth measurements of a thin object are made from opposite directions. Then the surface vanishes since the linear TSDF updates cancel each other out.
\mbox{\textbf{4-)}} All measurements are treated equally - independent of the direction they have been acquired. This assumption is usually incorrect since the noise level along the viewing direction is typically very different from the one in orthogonal directions.
\mbox{\textbf{5-)}} The fusion approach is unable to handle gross outliers. The depth map has to be pre-filtered or incorrect measurements will clutter the scene.
\mbox{\textbf{6-)}} The fusion parameters must be tuned for specific scenes and sensors and it is often difficult to find a good trade-off between runtime and different aspects of reconstruction quality.

In this paper, we aim to tackle the above mentioned disadvantages while maintaining all the advantages of traditional approach with a reasonable amount of additional computation time to still meet real-time requirements.
To this end, we propose a learned approach that fuses noisy and outlier contaminated measurements into a single surface, performs non-linear updates to better deal with object boundaries and thin structures, and is fast enough for real-time applications.
Figure~\ref{fig:teaser} shows example outputs of our approach.
%
In sum, this paper's \textbf{contributions} are as follows:
\begin{itemize}[topsep=2pt,leftmargin=*]
\setlength\itemsep{-1.3mm}
\item We present a learning-based method for real-time depth map fusion. 
Due to its compact architecture it requires only little training data, and is not prone to over-fitting.
\item We propose a scalable and real-time capable neural architecture that is independent of the scene size. Therefore, it is applicable to a large set of real-world scenarios.
\item We show significant improvement of standard TSDF fusion's shortcomings:
1) It better handles the fusion of anisotropic noise distributions that naturally arise from the multi-view setting, and
2) It mitigates the surface thickening effect on thin objects and surface boundaries by avoiding inconsistent updates.
\end{itemize}

\section{Related Work}

\boldparagraph{Volumetric Depth Map Fusion.}
With their seminal work, Curless and Levoy~\cite{Curless-Levoy-SIGGRAPH-1996} proposed an elegant way for fusing noisy depth maps which later got adopted by numerous works like KinectFusion~\cite{Izadi-et-al-SIGGRAPH-2011}, more scalable generalizations like voxel hashing~\cite{Niessner-et-al-SIGGRAPH-2013,Marniok-Goldluecke-WACV-2018}, or hierarchical scene representations, such as voxel octrees~\cite{Fuhrmann-Goesele-TOG-2011,Steinbruecker-et-al-ICCV-2013,Marniok-et-al-GCPR-2017} and hierarchical hashing~\cite{Kaehler-et-al-RAL-2016}.
%
Especially for SLAM pipelines like InfiniTAM~\cite{Kaehler-et-al-TVCG-2015}, volumetric fusion became a standard approach due to its real-time capability.
In this context, it was also extended to become more accurate and robust~\cite{Choi-et-al-CVPR-2015} as well as improve SLAM with additional surface registration of scene parts to account for pose drift as proposed in~\cite{Whelan-et-al-IJRR-2016, maier2017efficient, Dai-et-et-al-SIGGRAPH-2017}.
%
Approaches with additional median filtering~\cite{Rothermel-et-al-ISPRS-2016,Marniok-et-al-GCPR-2017,Marniok-Goldluecke-WACV-2018} improve the robustness and are still real-time capable but with limited effectiveness.
Global optimization approaches \cite{Zach-et-al-ICCV-2007,Kolev-et-al-IJCV-2009} even better deal with noise and outliers if they further leverage semantic information \cite{Haene-et-al-CVPR-2013,Cherabier-et-al-3DV-2016,Haene-et-al-TPAMI-2017,Savinov-et-al-CVPR-2015,Savinov-et-al-CVPR-2016}, but they are computationally expensive and not real-time capable.
In~\cite{zollhoefer2015shading, maier2017intrinsic3d}, the authors propose methods for refinement of already fused SDF geometry based on shape-from-shading.
The vast majority of these approaches directly fuse RGB-D images for which Zollh\"ofer \etal~\cite{Zollhoefer-et-al-CGF-STAR-2018} provide a recent survey.
All these methods handle noisy measurements by updating a wider band of voxels around the measured depth leading to thickening artifacts on thin geometry.

\boldparagraph{Surfel-based Fusion Methods.}
Surfel-based methods approximate the surface with local point samples, which can further encode additional local properties such as normal or texture information.
Multiple methods have been proposed, \eg MRSMap~\cite{Stueckler-Behnke-JVCIR-2014} uses an octree to store multi-resolution surfel data.
The point-based fusion methods~\cite{Keller-et-al-3DV-2013,Lefloch-et-al-FUSION-2015} combine a surfel representation with probabilistic fusion discussed in the next paragraph.
ElasticFusion~\cite{Whelan-et-al-IJRR-2016} handles real-time loop closures and corrects all surface estimates online.
Sch\"ops \etal~\cite{Schoeps-et-al-TPAMI-2019} proposed a depth fusion approach with real-time mesh construction.
A disadvantage of surfel-based methods is the missing connectivity information among surfels.
The unstructured neighborhood relationships can only be established with a nearest neighbor search or simplified with space partitioning data structures.
In our work, we decided to rely on volumetric representation, but extending our approach to unstructured settings is an interesting avenue of future work.

\boldparagraph{Probabilistic Depth Map Fusion.}
To account for varying noise levels in the input depth maps and along different line-of-sight directions, the fusion problem can also be cast as probability density estimation~\cite{Duan-et-al-ICPR-2012} while typically assuming a Gaussian noise model.
Keller~\etal~\cite{Keller-et-al-3DV-2013} propose a point-based fusion approach which directly updates a point cloud rather than a voxel grid.
Lefloch~\etal~\cite{Lefloch-et-al-FUSION-2015} extended this idea to anisotropic point-based fusion in order to account for different noise levels when a surface is observed from different incident angles.
The mesh-based fusion approach by Zienkiewicz \etal~\cite{Zienkiewicz-et-al-3DV-2016} allows for depth fusion across various mesh resolutions for known fixed topology.
The probabilistic fusion method by Woodford and Vogiatzis~\cite{Woodford-Vogiatzis-ECCV-2012} incorporates long range visibility constraints.
Similar ray-based visibility constraints were also used in \cite{Ulusoy-et-al-3DV-2015,Ulusoy-et-al-CVPR-2016}, but these methods are not real-time capable due to the complex optimization of ray potentials.
Anisotropic depth map fusion methods additionally keep track of fusion covariances~\cite{Ylimaeki-et-al-ICPR-2018}.
Similarly, PSDF Fusion~\cite{Dong-et-al-ECCV-2018} explicitly models directional dependent sensor noise.
In contrast to our method, all these approaches assume particular noise distributions, primarily Gaussians, which often do not model the real sensor observations correctly.
%

%

\begin{figure*}[t!]
	\vspace{-6pt}
	\includegraphics[width=\textwidth]{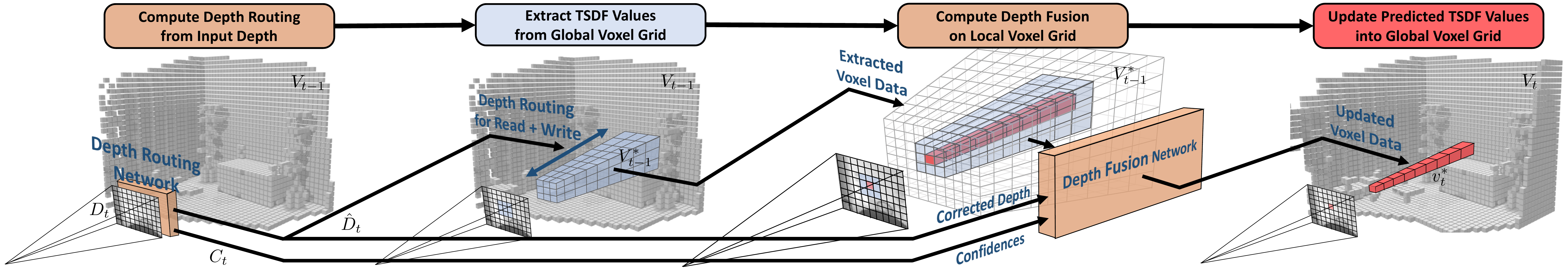}
	\vspace{-0.6cm}
	\caption{\textbf{System overview for integrating depth maps into a global TSDF volume.}
	A 2D \textbf{\textit{Depth Routing Network}} takes depth input and decides the update location for every ray within the TSDF volume.
	The network corrects for noise, outliers and missing values and further estimates per-ray confidence values.
	Then, for each ray, we extract a depth and view-dependent local voxel grid (light blue) which also includes neighboring rays. We sample $S$ values along each ray, centered around the surface.
	A \textbf{\textit{Depth Fusion Network}} then takes the local grid of existing TSDF values, the depth and confidences to predict adequate updates. The predicted TSDF values (red) are then written back into the global volume.
	Our method learns a robust weighting of input depths and performs non-linear updates to better handle noise, outliers, and thin objects.}
	\label{fig:system_overview}
	\vspace{-0.4cm}
\end{figure*}

\boldparagraph{Learning-based Reconstruction Approaches.}
Several learning-based methods have been proposed to fuse, estimate, or improve geometry.
SurfaceNet~\cite{Ji-et-al-ICCV-2017} jointly estimates multi-view stereo depth maps and their volumetric fusion, but is extremely memory demanding as each camera view requires a full voxel grid.
In \cite{Leroy-et-al-ECCV-2018}, multi-view consistency is learned upon classical TSDF fusion.
RayNet~\cite{Paschalidou-et-al-CVPR-2018} models view dependencies along ray potentials with a  Markov random field which is jointly learned with a view-invariant feature representation.
3DMV~\cite{Dai-Niessner-ECCV-2018} combines 2D view information with a pre-fused TSDF scene to jointly optimize for shape and semantics.
Riegler \etal~\cite{Riegler-et-al-3DV-2018} fuse depth maps using standard TSDF and subsequently post-process the fused model with a neural network.
Moreover, hierarchical volumetric deep learning-based approaches~\cite{Cao-et-al-ECCV-2018,Cherabier-et-al-ECCV-2018,Dai-et-al-CVPR-2018} tackle the effects of noisy measurements, outliers, and missing data.
All these approaches operate on a voxel grid with high memory demands and are not real-time capable.
Further, there are several works that learn to predict 3D meshes based on input images~\cite{Groueix-et-al-CVPR-2018, Gkioxari-et-al-ICCV-2019, Wen-et-al-ICCV-2019}.

\boldparagraph{Learned Scene Representations.}
Ladicky \etal~\cite{Ladicky-et-al-ICCV-2017} directly estimate an iso-surface from a point cloud via learned local point features with a random forest.
In addition, there exist multiple proposals for methods that learn 3D reconstruction in an implicit space~\cite{Mescheder-et-al-CVPR-2019, Park-et-al-CVPR-2019, Chen-et-al-CVPR-2019, Michalkiewicz-et-al-ICCV-2019}. 
These methods show promising results, but they operate only on a unit cube and are thus limited to single objects or small scenes and they are neither suited for online reconstruction.
Contrary to all these methods, our method is independent of the scene's size and can thus also operate on large-scale scenes. 
Additionally, our method uses learning in an online process, which allows to leverage already fused information for fusing a new depth map. 
In~\cite{Bloesch-et-al-CVPR-2018, Zhi-et-al-CVPR-2019}, the authors propose neural models that learn a compact and optimizable 2.5D depth representation for SLAM applications.
DeepTAM~\cite{Zhou-et-al-ECCV-2018} also addresses SLAM, but the mapping part heavily relies on hand-crafted photoconsistencies and corresponding weights to form a traditional cost volume for depth estimation.
None of these methods address global model fusion.

\section{Method}
\label{sec:method}

We first review the standard TSDF fusion approach to provide context and to introduce notation before we present our learned TSDF fusion method.

\subsection{Review of Standard TSDF Fusion}
Standard TSDF fusion integrates given depth maps $\boldsymbol{D}_{t = 1, \dots, T} \in \mathds{R}^{W \times H}$ from known viewpoints $\boldsymbol{P}_t \in \text{SE}(3)$ with  camera intrinsics $\boldsymbol{K}_t$ into a discretized signed distance function $\boldsymbol{V}_t \in \mathds{R}^{X \times Y \times Z}$ and weight function $\boldsymbol{W}_t \in \mathds{R}^{X \times Y \times Z}$ defined over the entire scene.
The fusion process is incremental, \ie each depth map is integrated after one another for location $\boldsymbol{x}$ using the update equations introduced by Curless and Levoy~\cite{Curless-Levoy-SIGGRAPH-1996} as
\begin{align}
\boldsymbol{V}_{t}(\boldsymbol{x}) &= \frac{\boldsymbol{W}_{t-1}(\boldsymbol{x}) \cdot \boldsymbol{V}_{t-1}(\boldsymbol{x}) + w_{t}(\boldsymbol{x}) \cdot v_t(\boldsymbol{x})}{\boldsymbol{W}_{t-1}(\boldsymbol{x}) + w_{t}(\boldsymbol{x})}  \label{eq:tsdf-update1} \\
\boldsymbol{W}_t(\boldsymbol{x}) &= \boldsymbol{W}_{t-1}(\boldsymbol{x}) + w_{t}(\boldsymbol{x}) \enspace ,
\label{eq:tsdf-update2}
\end{align}
starting from zero-initialized volumes $\boldsymbol{V}_0$ and $\boldsymbol{W}_0$.
The signed distance update $v_t$ and its corresponding weight $w_t$ integrate the depth measurements of the next depth map $\boldsymbol{D}_t$ at time step $t$ into the TSDF volume.
These update functions are traditionally truncated before and after the surface in order to ensure efficient runtimes and robust reconstruction of fine-structured surfaces given noisy depth measurements.

The choice of the truncation distance parameter typically requires cumbersome hand-tuning to adapt to a specific scene and depth sensor as well as accounting for runtime.
If the truncation distance is chosen too large, the reconstruction of thin structures becomes more difficult due to larger thickening artifacts and the fusion process gets slower since more voxels have to be updated for each ray.
Contrary, a small truncation distance results in time efficient updates but cannot deal with larger noise in the depth measurements.

In this paper, we overcome this limitation by learning the function $v_t$ automatically from data.
Our system is based on the same above mentioned update equations and our learned functions have only little computational overhead compared to traditional TSDF fusion.
As such, our method facilitates real-time depth map fusion and can be readily integrated into existing reconstruction systems.
In the following, we describe our proposed method in more detail.

\subsection{System Overview}
Our method contains \textit{two} network components: a \textit{\textbf{depth routing network}} and a \textit{\textbf{depth fusion network}}. 
The pipeline consists of the following \textit{four} essential processing steps which are also illustrated in Figure~\ref{fig:system_overview}:
\begin{enumerate}[topsep=1pt,leftmargin=*]
	\setlength\itemsep{-1mm}
	\item \textbf{Depth Routing:} The depth routing network takes a raw depth map $\boldsymbol{D}_t$ and estimates a denoised and outlier-corrected depth map $\boldsymbol{\hat{D}}_t$, and further estimates a corresponding confidence map $\boldsymbol{C}_t$.
	This network \textit{routes} the depth location for reading and writing TSDF values along each viewing ray.
	\item \textbf{TSDF Extraction:} Given the routed depth values $\boldsymbol{\hat{D}}_t$, we extract a local camera-aligned voxel grid with TSDF data $\boldsymbol{V}^*_{t-1}$ and weight $\boldsymbol{W}^*_{t-1}$ via trilinear interpolation from the corresponding global voxel grids $\boldsymbol{V}_{t-1}$, $\boldsymbol{W}_{t-1}$. 
	\item \textbf{Depth Fusion:} The depth fusion network takes the results of the previous processing steps $(\boldsymbol{\hat{D}}_t, \boldsymbol{C}_t, \boldsymbol{W}^*_{t-1}, \boldsymbol{V}^*_{t-1})$ and computes the local TSDF update $v^*_t$.
	\item \textbf{TSDF Update Integration:} The predicted TSDF update $v^*_t$ is transferred back into the global coordinate frame to get $v_t$ which is then integrated into the global TSDF volumes $\boldsymbol{V}_{t}, \boldsymbol{W}_{t}$ using the TSDF updates in Eqs.~\eqref{eq:tsdf-update1},~\eqref{eq:tsdf-update2}.
\end{enumerate}
%
%

%
\noindent
These processing steps are detailed in the next subsections.

\begin{figure*}[t]
	\vspace{-0.2cm}
	\includegraphics[width=\linewidth]{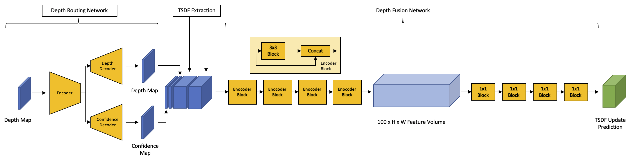}
	\vspace{-0.7cm}
	\caption{{\bf Proposed network architecture.} Our depth routing network consists of a U-Net (depth one) with two separate decoders predicting a corrected depth map and a corresponding confidence map. The depth fusion network extracts in a series of encoding blocks 100 features along each ray. These features are then used to predict the TSDF updates along the ray. }	
	\label{fig:network_architecture}
	\vspace{-0.4cm}
\end{figure*}

\subsection{Depth Routing}
Using the depth routing network, we pre-process the depth maps before passing them to the depth fusion network with the main motivation of denoising and outlier correction.
Towards this end, the network predicts denoised depth maps and also per-pixel confidence maps $\boldsymbol{C}_{t = 1, \dots, T} \in \mathds{R}^{W \times H}$.
Figure \ref{fig:network_architecture} illustrates our network architecture,
which is using a fully-convolutional U-Net~\cite{Ronneberger-et-al-MICCAI-2015} with a joint encoder and separate decoders for confidence and depth prediction.
Further, we do not use normalization layers since it negatively influences the depth prediction performance by adding a depth-dependent bias to the result.
The depth map and the confidence map are processed by two separate decoders to which the output of the bottleneck layers serves as an input.

\subsection{TSDF Extraction}  \label{sec:tsdf-extraction}
%
Instead of processing each ray of a view $t$ independently as in standard TSDF fusion, we deliberately choose to compute the TSDF updates based on the data of a larger neighborhood in order to make a more informed decision about the surface location.
Further, the 2D input data also holds valuable information about surface locations as often indicated by depth discontinuities.
%
%
%
We argue that the fusion network can best benefit from both 2D and 3D data sources when they are already in correspondence and therefore propose a view-aligned local neighborhood extraction.
Then, the 3D TSDF data and the 2D input data can be easily concatenated and fed into the network.
Hence, for efficient real-time updates of the global data $\boldsymbol{V}_{t-1}$, $\boldsymbol{W}_{t-1}$, we extract a local, view-dependent TSDF volume and corresponding weights $\boldsymbol{V}^*_{t-1}, \boldsymbol{W}^*_{t-1} \in \mathds{R}^{W \times H \times S}$.
%
The first two volume dimensions $W, H$ correspond to the width and height of the depth map whereas the third dimension $S$ represents the local depth-sampling dimension of the window sampled along the ray. 
This number $S$ closely relates to the truncation distance in standard TSDF fusion.
For each ray independently, the local windows are centered at their respective depth values $\boldsymbol{\hat{D}}_t$ and discretely sampled into a fixed number of $S$ values from the volume $\boldsymbol{V}_{t-1}$.
We choose the step size of the sampling according to the resolution of the scene and use trilinear interpolation to mitigate sampling artifacts.

The input $\boldsymbol{I}_t$ to the subsequent depth fusion is then a combination of all available local information, that is,
corrected depth map $\boldsymbol{\hat{D}}_t$, confidence map $\boldsymbol{C}_t$ as well as the extracted TSDF values $\boldsymbol{V}^*_{t-1}$ and TSDF weights $\boldsymbol{W}^*_{t-1}$
\begin{equation}
\boldsymbol{I}_t = \begin{bmatrix} \boldsymbol{\hat{D}}_t &  \boldsymbol{C}_t & \boldsymbol{W}^*_{t-1} & \boldsymbol{V}^*_{t-1} \end{bmatrix} \in \mathds{R}^{W \times H \times (2S + 2)} \enspace .
\end{equation}
Before the subsequent update prediction step, we explicitly filter gross outliers where $\boldsymbol{C}_{t} < C_\text{thr}$ and set their corresponding feature values in $\boldsymbol{I}_t$ to zero. 

\subsection{Depth Fusion} \label{sec:depth-fusion}
%
Our depth fusion network takes the local 3D feature volume $\boldsymbol{I}_t$ as input and predicts the local TSDF update $v^*_t \!\in\! \mathds{R}^{W \times H \times S}$.
The architecture is fully convolutional in two dimensions and the channel dimension is along the camera viewing direction. 
Our network is relatively compact and thereby facilitates real-time computation.

Our depth fusion network operates in a two-stage approach, as shown in Figure~\ref{fig:network_architecture}.
The first stage encodes local and global information in the viewing frustum.
We sequentially pass the input 3D feature volume through encoding blocks of two consecutive convolutional layers with interleaved batch normalization, non-linear activation using leaky ReLUs, and a dropout layer. 
The output of every block is concatenated with its input and passed through the next block. 
With every block, the receptive field of the neural network increases.
This sequential feature extraction results in a 100-dimensional feature vector for each ray in the viewing frustum. 

The second network part takes the feature volume and predicts the TSDF updates along each ray.
The number of features is sequentially reduced by passing them through convolutional blocks with two $1 \times 1$ convolutional layers interleaved with leaky ReLUs, batch normalization, and dropout layers.
In the last block, we directly reduce from 40 features to 20 in the first layer and then to $S$ TSDF values in the last convolutional layer, where we apply a tanh-activation on the output mapping it to the range $[-1,1]$.

Note that predicted TSDF update values $\boldsymbol{v}_t^*$ can take any value. 
The network can decide to not update the TSDF at all, \eg, in case of an outlier.
Conversely, it can reduce the influence of existing TSDF values if they contained outliers.

%
%



\subsection{TSDF Update Integration} \label{sec:update-integration}
%
In order to compute the updated global TSDF volume $\boldsymbol{V}_t$ we transform the predicted local TSDF updates $v^*_t$ back into the global coordinate frame $v_t$.
To this end, we apply the inverse operation of the previous extraction step, that is,
we redistribute the values using the same trilinear interpolation weights.
%
In fact, we actually repurpose the update weights $w_t$ for this task, where $\boldsymbol{W}_t$ accumulates the splatting weights for each voxel in the scene.
Moreover, we also use $\boldsymbol{W}_t$ for post-filtering extreme outliers\footnote{See supplementary material for further information.}.


\subsection{Loss Function and Training Procedure}
The two networks in our pipeline are trained in two steps.
First, we train the depth routing network and then use the pre-trained routing output to train the fusion network.


\boldparagraph{Depth Routing Network.}
%
We train the depth prediction head in a supervised manner by computing the L1 loss on absolute depth values as well as on the depth map gradient, as proposed in~\cite{Donne-et-al-CVPR-2019}. 
For training the confidence head, we chose a self-supervised approach~\cite{Kendall-et-al-NIPS-2017}. 
Therefore, the final loss function has the form
\begin{equation}
  \mathcal{L}_{\text{2D}} = \sum_i c_i \mathcal{L}_1(y_i, \hat{y}_i) +  c_i \mathcal{L}_1(\nabla y_i, \nabla \hat{y}_i) - \lambda \log{c_i}
\end{equation}

where $y_i, \hat{y}_i$ are the predicted and ground-truth depth values at pixel $i$ respectively and $c_i \in \boldsymbol{C}_t$ is the confidence value. 
The hyperparameter $\lambda$ is empirically set to $0.015$.

\boldparagraph{Depth Fusion Network.}
%
Despite the pre-processing of the routing network, the filtered depth map might still contain noise and outliers which should be further handled by the depth fusion network.
Each global TSDF update step should a) integrate new information about the true geometry and b) not destroy valuable, previously fused surface information.
We train the fusion network in a supervised manner by choosing random update steps at time $t$ during the fusion and penalize differences between the updated local volume $\boldsymbol{V}^*_t = \frac{\boldsymbol{W}^*_{t-1} \cdot \boldsymbol{V}^*_{t-1} + w^*_t \cdot v^*_t}{\boldsymbol{W}^*_{t-1} + w^*_t} \!\in\! \mathds{R}^{W \times H \times S}$ and the local ground-truth $\boldsymbol{\hat{V}}^* \!\in\! \mathds{R}^{W \times H \times S}$. Therefore, we define the loss function over all rays $i$ as
\begin{equation}
\mathcal{L}_{\text{3D}} = \sum_{i}{\lambda_1 \mathcal{L}_{1}(\boldsymbol{V}^*_{ti}, \boldsymbol{\hat{V}}_i^*) + \lambda_C D_C(\boldsymbol{V}^*_{ti}, \boldsymbol{\hat{V}}_i^*)} 
\end{equation}
Here, $\mathcal{L}_1$ denotes the L1 loss over raw TSDF values and $D_C$ denotes the cosine distance between the signs of the TSDF values computed along each ray $i$. 
The goal of the first term is to preserve fine surface detail (through means of $\mathcal{L}_1$), while, the term $D_C$ ensures that the surface is located at the zero-crossing of the signed distance field.
The weights $\lambda_1 = 1$ and $\lambda_C = 0.1$ have been empirically found.
\section{Experiments}

In this section, we first present additional implementation details and our experimental setup.
Next, we evaluate and discuss the efficacy of our approach on both synthetic and real-world data. 
We demonstrate that our approach outperforms traditional TSDF fusion and state-of-the-art learning-based approaches in terms of reconstruction accuracy with only little computational overhead.

\subsection{Implementation Details}

All networks were implemented in PyTorch and trained on an NVIDIA TITAN Xp GPU. 
We trained both networks using the RMSProp optimization algorithm with momentum $0.9$ and initial learning rate $1\mathrm{e}{-5}$ for the depth routing network and $1\mathrm{e}{-3}$ for the depth fusion network.
The dropout layers were set to a probability of $0.2$.
For all experiments, we trained our neural networks in a sequential process, where we first pre-trained the depth routing and then the depth fusion network.
A joint end-to-end refinement did not lead to an improvement of the overall performance of the system.
To train the depth routing network, we use 10K frames sampled from 100 ModelNet~\cite{Modelnet-CVPR-2015} or ShapeNet~\cite{shapenet2015} objects and perturb them with artificial speckle noise. 
The data is packed into batches of size 4 and the gradient is accumulated across 8 batches before updating the routing network weights. 
Because of the incremental nature of the TSDF update equation, we must train our depth fusion network using a batch size of $1$. 
However, each batch updates a very large number of voxels in the volume over which the loss is defined and, together with batch normalization, we obtain robust convergence during training.
Since our network has only very few parameters, it is hard to overfit and only little training data is required.
In fact, we can train our entire network (given a pre-trained depth routing network) on only ten models from ModelNet~\cite{Modelnet-CVPR-2015} or ShapeNet~\cite{shapenet2015} with a total of 1000 depth maps and it already generalizes robustly to other scenes.
Furthermore, we can train the network from scratch in only 20 epochs (each epoch passes once over all 1000 frames).
Unless otherwise specified, we used $S=9$ and $C_{thr}=0.9$ across all experiments\footnote{See supplementary material for further evaluation.}.
For all experiments, we used a voxel size $0.008 m$, corresponding to a grid resolution of $128^3$ for ShapeNet and ModelNet.


\boldparagraph{Runtime.} A forward pass through the depth routing network and the depth fusion network for one depth map ($W=320, H=240$) takes 0.9 ms and 1.8 ms, respectively while the full pipeline runs at 15 fps.
These numbers can be improved with a more efficient implementation, but already meet real-time requirements. 

\subsection{Results}
%
We evaluate our method on synthetic and real-world data comparing to traditional TSDF fusion~\cite{Curless-Levoy-SIGGRAPH-1996} as a baseline as well as to the state-of-the-art PSDF fusion method presented by Dong \etal~\cite{Dong-et-al-ECCV-2018}. 
Moreover, we compare to state-of-the-art learning-based 3D reconstruction methods OccupancyNetworks~\cite{Mescheder-et-al-CVPR-2019} and DeepSDF~\cite{Park-et-al-CVPR-2019}.

\boldparagraph{Evaluation Metrics.} 
For quantifying the performance of our method, we compute the following four metrics by comparing the estimated TSDF against the ground-truth.
\begin{itemize}[topsep=1pt,leftmargin=*]
	\setlength\itemsep{-1mm}
	\item \textbf{MAD:} The mean absolute distance is computed over all TSDF voxels and measures the reconstruction performance on fine surface details.
	\item \textbf{MSE:} The mean squared error loss is computed over all TSDF voxels and measures the reconstruction performance on large surface deviations.
	\item \textbf{Accuracy:} 
	We compare the actual reconstruction accuracy on the occupancy grid. We extract the occupancy grid in the ground-truth and the estimated TSDF by extracting all voxels with negative TSDF values.
	\item \textbf{Intersection over Union (IoU):} We compute the intersection-over-union on the occupancy grid, which is an alternative performance measure to the accuracy.
\end{itemize}
These metrics not only quantify how well our pipeline fuses depth maps into a TSDF, but also how well it performs in classifying the occupancy and reconstructing the geometry.

\subsection{Synthetic Data}
%
To evaluate our method's performance in fusing noisy synthetic data, we train and test it on the ModelNet~\cite{Modelnet-CVPR-2015} and ShapeNet~\cite{shapenet2015} datasets using rendered ground-truth depth maps that are perturbed with an artificial depth-dependent multiplicative noise distribution. 
For both, ModelNet and ShapeNet, we randomly sample our training and test data from the official train-test split.

\begin{table}[t]
	\centering
	\scriptsize
	\setlength{\tabcolsep}{3.5mm} 
	\begin{tabular}{lrlcc}
		\toprule
		\multirow{2}{*}{\bf Method} & {\bf MSE} & {\bf MAD} & {\bf Acc.} & {\bf IoU} \\ 
		& {\scriptsize [e-05]} &    & {\scriptsize [\%]} & {\scriptsize [0, 1]} \\
		\midrule
		DeepSDF~\cite{Park-et-al-CVPR-2019} & 464.0 & 0.0499 & 66.48 & 0.538\\
		OccupancyNetworks~\cite{Mescheder-et-al-CVPR-2019} & 56.8 & 0.0166 & 85.66 & 0.484 \\[2pt]
		TSDF Fusion~\cite{Curless-Levoy-SIGGRAPH-1996} & 11.0 & 0.0078 & 88.06 & 0.659 \\
		TSDF Fusion + Routing & 27.0 & 0.0084 & 87.48 & 0.650 \\[2pt]
		Ours w/o Routing & \textbf{5.9} & 0.0051 & 93.91 & 0.765  \\
		Ours &  \textbf{5.9} & \textbf{0.0050} & \textbf{94.77} & \textbf{0.785} \\
		\bottomrule 
	\end{tabular}
	\vspace{-5pt}
	\caption{\textbf{Quantitative Results on ShapeNet~\cite{shapenet2015}}. 
		Our method outperforms TSDF fusion and other learning-based approaches in fusing noisy ($\sigma = 0.005$) depth-maps rendered from ShapeNet objects. 
		The benefit of the routing network increases with higher noise levels (see Fig.~\ref{fig:noise}).}
	\label{tab:shapenet-comparison}
	\vspace{-8pt}
\end{table}

\begin{figure}[t]
	\vspace{-4pt}
	\scriptsize
	\centering
	\setlength{\tabcolsep}{0.10mm}
	\begin{tabular}{ccccc}
		\includegraphics[width=0.2\columnwidth,trim={0 0 0 35},clip]{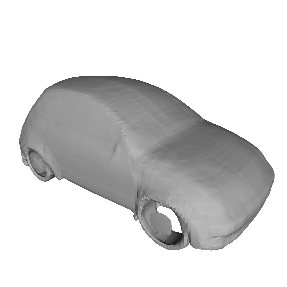} &
		\includegraphics[width=0.2\columnwidth,trim={0 0 0 35},clip]{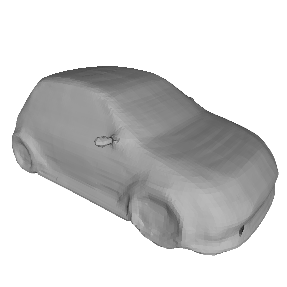} &
		\includegraphics[width=0.2\columnwidth,trim={0 0 0 35},clip]{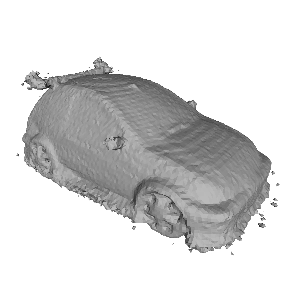} &
		\includegraphics[width=0.2\columnwidth,trim={0 0 0 35},clip]{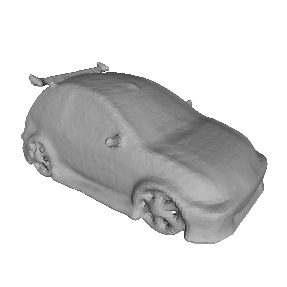} &
		\includegraphics[width=0.2\columnwidth,trim={0 0 0 35},clip]{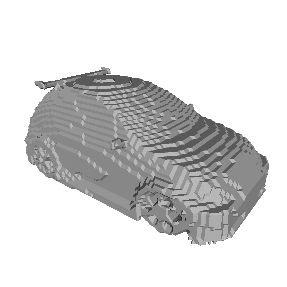} \\[-1mm]
		\includegraphics[width=0.2\columnwidth,trim={37 45 37 62},clip]{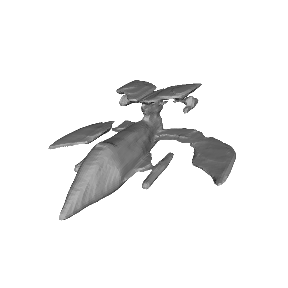} &
		\includegraphics[width=0.2\columnwidth,trim={37 45 37 62},clip]{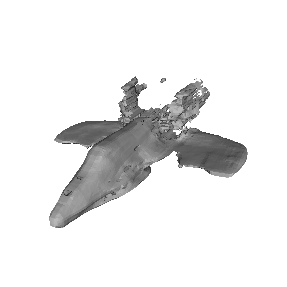} &
		\includegraphics[width=0.2\columnwidth,trim={37 45 37 62},clip]{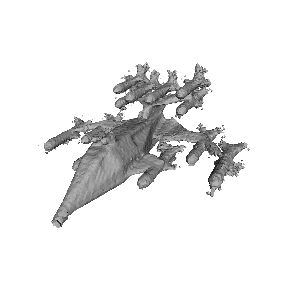} &
		\includegraphics[width=0.2\columnwidth,trim={37 45 37 62},clip]{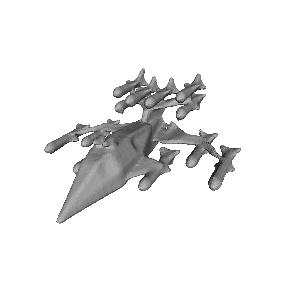} &
		\includegraphics[width=0.2\columnwidth,trim={37 45 37 62},clip]{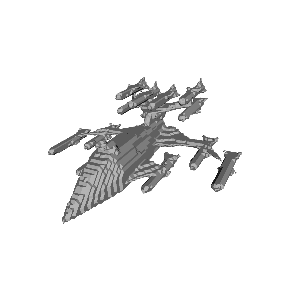} \\[-1mm]
		\includegraphics[width=0.2\columnwidth,trim={55 10 55 20},clip]{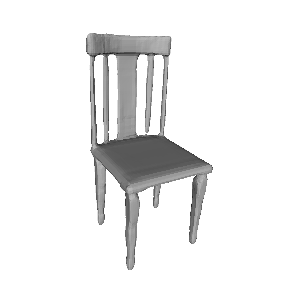} &
		\includegraphics[width=0.2\columnwidth,trim={45 2 45 12},clip]{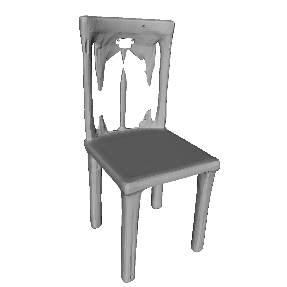} &
		\includegraphics[width=0.2\columnwidth,trim={45 2 45 12},clip]{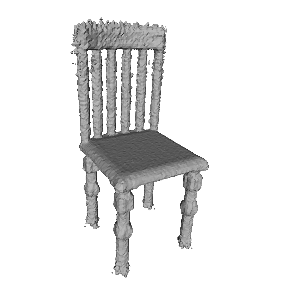} &
		\includegraphics[width=0.2\columnwidth,trim={45 2 45 12},clip]{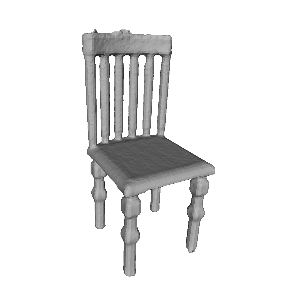} &
		\includegraphics[width=0.2\columnwidth,trim={45 2 45 12},clip]{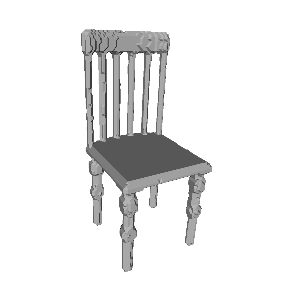} \\
		\sfc{DeepSDF~\cite{Park-et-al-CVPR-2019}} & \sfc{Occ.Net.~\cite{Mescheder-et-al-CVPR-2019}}
		& \sfc{TSDF~\cite{Curless-Levoy-SIGGRAPH-1996}} & \sfc{Ours} & \sfc{GT}\\[-6pt]
	\end{tabular}
	\caption{\textbf{Qualitative Results on ShapeNet~\cite{shapenet2015}}. Our method is superior to all other methods in reconstructing fine details (see car wheels and spoiler) and produces smoother surfaces (input noise level $\sigma = 0.005$).}
	\label{fig:qualitative-shapenet-results}
	\vspace{-16pt}
\end{figure}

\begin{figure}[t]
	\vspace{-2mm}
	\hspace{-2mm}
	\begin{subfigure}[c]{0.6\columnwidth}
		\centering
		\includegraphics[width=1.02\linewidth]{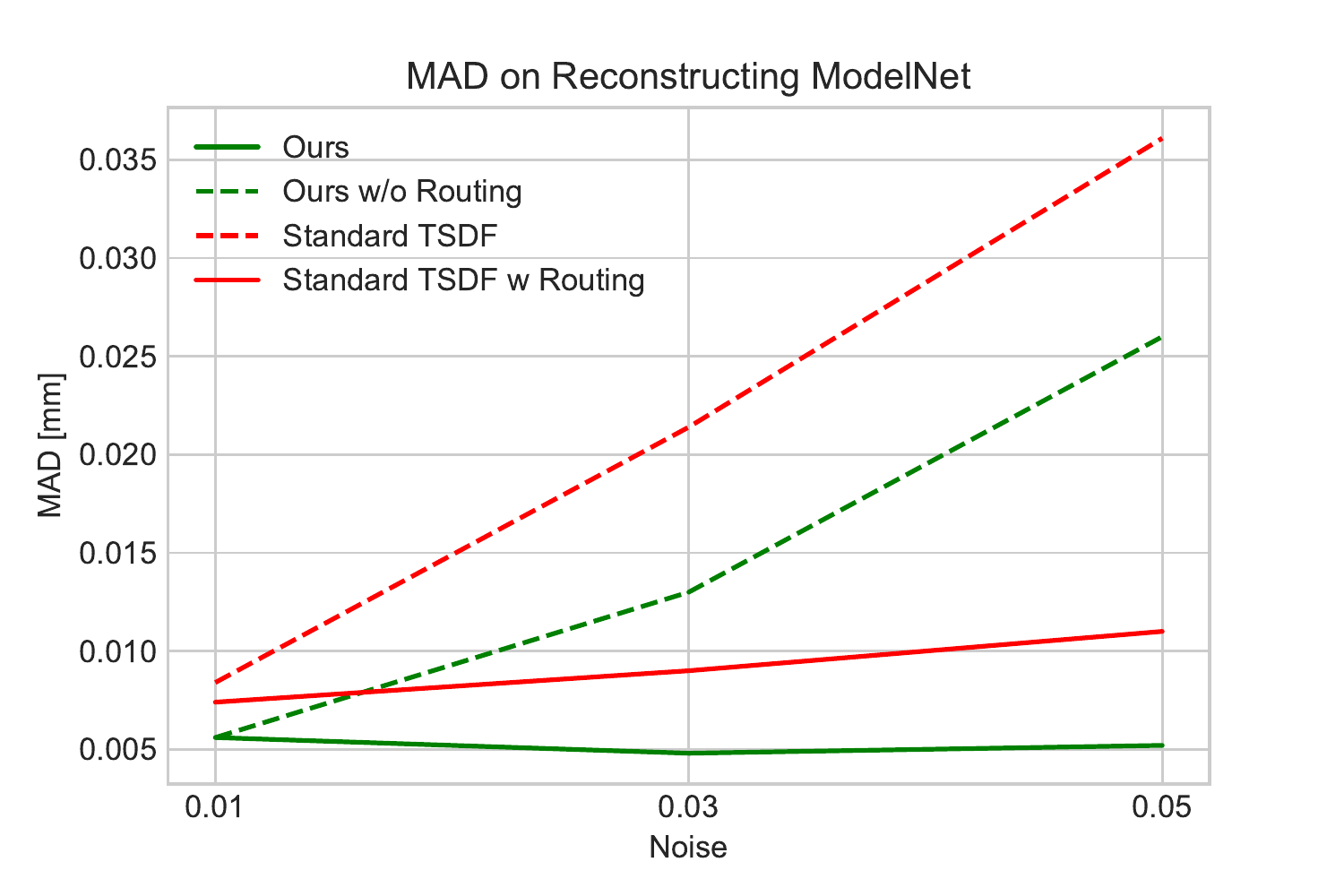}
	\end{subfigure}%
    \hspace{-6mm}
	\begin{subfigure}[c]{0.4\columnwidth}
		\centering
		\setlength{\tabcolsep}{0mm}
		\newcommand{\sz}{0.38}
		\vspace{-3pt}
		\begin{tabular}{cccc}
			\includegraphics[width=\sz\columnwidth,trim={17 0 30 0},clip]{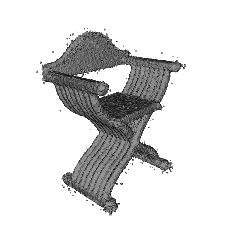} &
			\includegraphics[width=\sz\columnwidth,trim={15 0 27 0},clip]{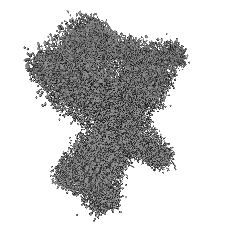} &
			\includegraphics[width=\sz\columnwidth,trim={12 0 15 0},clip]{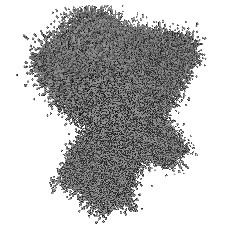} &
			\rotatebox{90}{\hspace{2pt} \sfc{\tiny Standard TSDF}}\\[-7pt]
			\includegraphics[width=\sz\columnwidth,trim={17 0 30 17},clip]{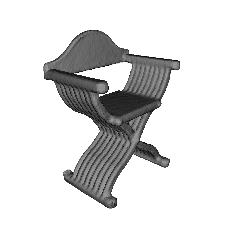} &
			\includegraphics[width=\sz\columnwidth,trim={15 0 27 17},clip]{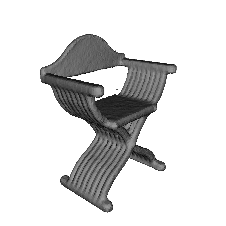} &
			\includegraphics[width=\sz\columnwidth,trim={12 0 15 17},clip]{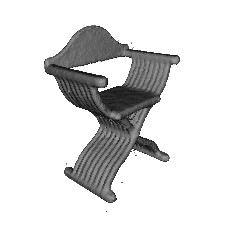} &
			\rotatebox{90}{\hspace{12pt} \sfc{\tiny Ours}}\\[-9pt]
			\tiny \sfc{0.01} & \tiny \sfc{0.03} & \tiny \sfc{0.05} & \\
		\end{tabular}		
	\end{subfigure}
	\vspace{-10pt}
\caption{\textbf{Evaluation of different noise levels $\sigma$}. 
	The left plot shows MAD for different noise levels $\sigma \in \{0.01, 0.03, 0.05\}$. 
	Our routing network stabilizes both, our method as well as standard TSDF fusion, for high noise levels. On the right, we show corresponding qualitative results on ModelNet test data for Standard TSDF and our method. 
	The figures show the denoising capability of our method, where standard TSDF fusion completely fails.}
	\label{fig:noise}
	\vspace{-15pt}
\end{figure}

\boldparagraph{ShapeNet.}
The model trained on ShapeNet is then used to evaluate the performance of our method in comparison with other approaches.
Therefore, we fuse noisy depth maps of 60 objects (10 per test class - plane, sofa, lamp, table, car, chair) from the test set, which have not been seen during training.
For comparison, we use the provided pre-trained model for point cloud completion in the case of OccupancyNetworks. 
In the case of DeepSDF, we trained the model from scratch using the code provided by the authors and using ShapeNet as training data.
The quantitative results of this evaluation are shown in Table~\ref{tab:shapenet-comparison}.
Our method consistently outperforms standard TSDF fusion as well as the pure learning-based approaches OccupancyNetworks~\cite{Mescheder-et-al-CVPR-2019} and DeepSDF~\cite{Park-et-al-CVPR-2019} on all metrics. 
Our method significantly improves the accuracy of the fused implicit mesh as well as their IoU, MAD and MSE scores. 
The results also indicate the potential of our routing network.
However, the full benefit of our routing network only becomes obvious when looking at the real-world data experiments and Figure~\ref{fig:noise}. 
%

Figure~\ref{fig:qualitative-shapenet-results} illustrates the strengths of our method in dealing with noise and in reconstructing thin structures. 
Flat surfaces in the ground-truth appear smoother in our results as compared to standard TSDF fusion. 
Furthermore, thin structures are better reconstructed and contain less thickening artifacts.
The thickening artifacts are also visible on the car's rims, where our method yields accurate results and DeepSDF and OccupancyNetworks both fail.
Both DeepSDF and OccupancyNetworks tend to over-smooth surface details less common in the training data, \eg, the spoiler of the car or the details on the chair's legs.

\boldparagraph{ModelNet.}
In order to test our method's robustness against noise we trained and evaluated on various noise levels $\sigma \in \{0.01, 0.03, 0.05\}$ and compared it to standard TSDF fusion.
We also analyze the effect of the depth routing network on the fusion result by omitting it in our pipeline and by testing it in combination with standard TSDF fusion.
%
%
%
%
Figure~\ref{fig:noise} illustrates that our pipeline outperforms standard TSDF fusion for all tested noise levels.
It also shows that our depth routing network stabilizes the fusion of data corrupted with extreme noise levels.
When used for data pre-processing, our depth routing network also improves the results of standard TSDF fusion.

%
%


\subsection{Real-World Data}
%
\begin{figure*}[tb]
	\small
	\centering
	\setlength{\tabcolsep}{2.6mm}
	\newcommand{\sz}{0.3}
	\begin{tabular}{cccccc}
		\includegraphics[width=0.25\columnwidth,height=3cm]{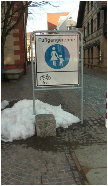} &     
		\includegraphics[width=\sz\columnwidth,height=3cm]{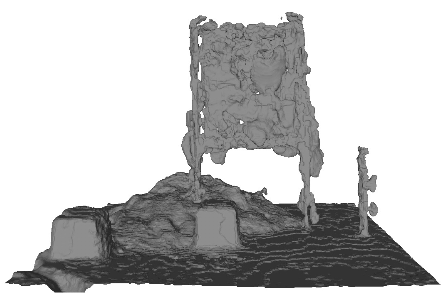} &
		\includegraphics[width=\sz\columnwidth,height=3cm]{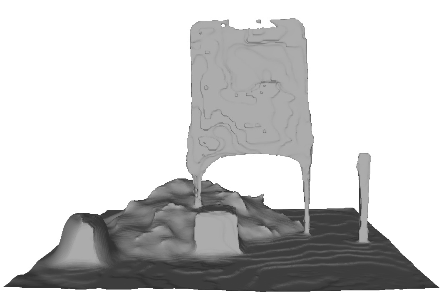} &     
		\includegraphics[width=\sz\columnwidth,height=3cm]{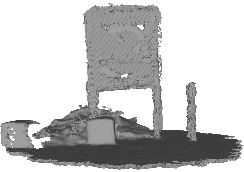} &  
		\includegraphics[width=\sz\columnwidth,height=3cm,trim={117 55 107 57},clip]{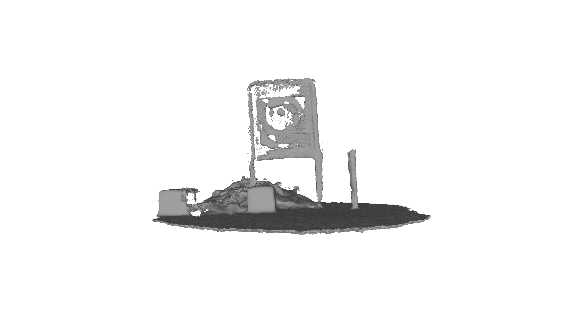} &
		\includegraphics[width=\sz\columnwidth,height=3cm,trim={117 55 107 57},clip]{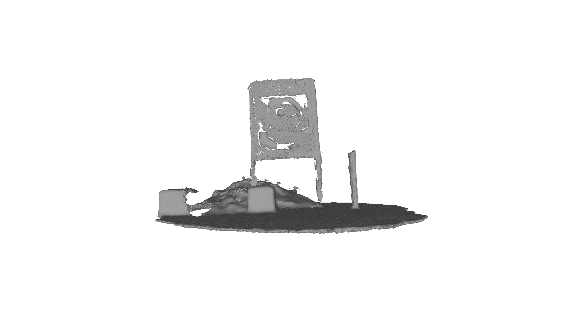} \\
		\includegraphics[width=0.25\columnwidth,height=3cm,trim={7 0 7 0},clip]{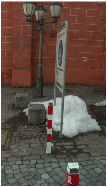} &     
		\includegraphics[width=\sz\columnwidth,height=3cm]{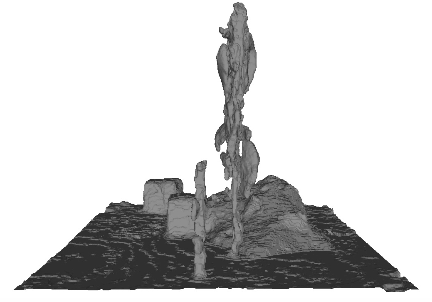} &
		\includegraphics[width=\sz\columnwidth,height=3cm]{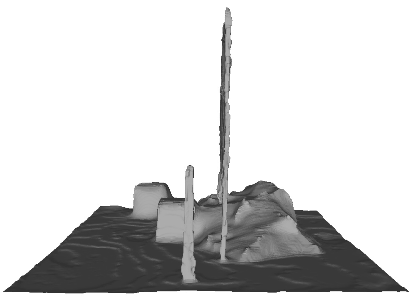} &     
		\includegraphics[width=\sz\columnwidth,height=3cm]{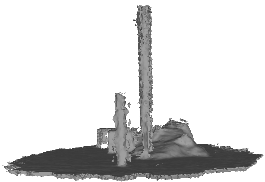} & 
		\includegraphics[width=\sz\columnwidth,height=3cm,trim={110 55 107 40},clip]{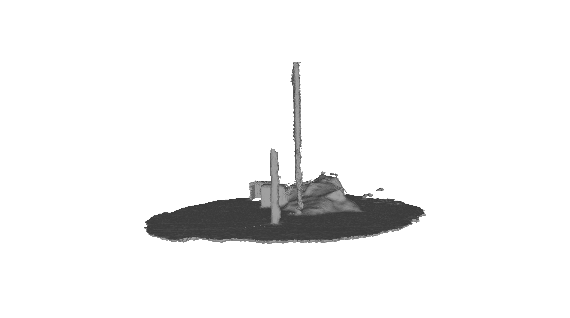} &
		\includegraphics[width=\sz\columnwidth,height=3cm,trim={110 55 107 40},clip]{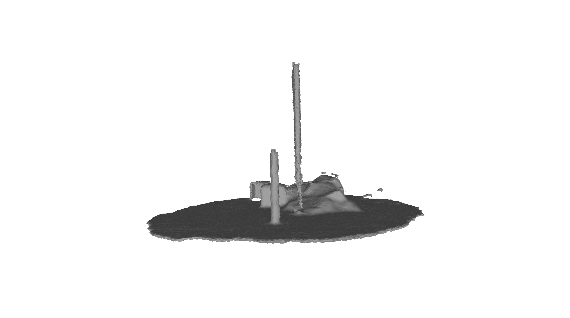} \\
		\sfc{RGB} & \sfc{TV-Flux~\cite{Zach-3DPVT-2008}} & \sfc{Ray Potentials~\cite{Savinov-et-al-CVPR-2016}} & \sfc{Standard TSDF~\cite{Curless-Levoy-SIGGRAPH-1996}} & \sfc{Ours w/o Routing} & \sfc{Ours}\\[-0.2cm]
	\end{tabular}
	\caption{\textbf{Qualitative results of our method on the Roadsign dataset~\cite{Ummenhofer-Brox-ICCV-2013}. } 
	Our method compares favorably to standard TSDF fusion as well as TV-Flux~\cite{Zach-3DPVT-2008} in reconstructing thin surfaces while showing comparable performance with ray potentials~\cite{Savinov-et-al-CVPR-2016}.
	Our method generalizes reasonably well since it was trained on ModelNet and never saw the noise and outlier statistics of stereo depth maps nor the shape statistics of this scene and therefore has a less complete output. ($C_{thr} = 0.5$)}
	\label{fig:street_sign_results}
\end{figure*}
%
We also evaluate on real-world datasets and compare to other state-of-the-art fusion methods. 
Due to lack of ground-truth data, we use the model trained on synthetic ModelNet data using an artificial and empirically chosen depth-dependent noise distribution with $\sigma = 0.01$. 
As such, we also show that our method must not necessarily be trained on real-world data but generalizes robustly to the real domain from being trained on noisy synthetic data only. 


%
%

\boldparagraph{3D Scene Data~\cite{Zhou-et-al-SIGGRAPH-2013}.}
To quantify the improvement of the reconstruction result, we evaluate our method compared to standard TSDF fusion on scenes provided by Zhou \etal~\cite{Zhou-et-al-SIGGRAPH-2013}. 
Since there is no volumetric ground-truth available for these scenes, we fuse all frames of each scene using standard TSDF fusion and denoised the meshes.
Then, we only fuse every 10th frame using standard TSDF fusion as well as our method for evaluation.

Table~\ref{tab:scene3d-comparison} shows the quantitative reconstruction results from fusing 5 scenes of the 3D scene dataset~\cite{Zhou-et-al-SIGGRAPH-2013}. 
Our method significantly outperforms standard TSDF fusion on all scenes without being trained on real-world data.

\begin{table}[t]
	\centering
	\scriptsize
	\setlength{\tabcolsep}{1.0mm}
	\begin{tabular}{lccccc}
		\toprule
		{\bf Method} & {\bf Lounge}  & {\bf Copyroom} & {\bf Stonewall} & {\bf Cactusgarden} & {\bf Burghers} \\ 
		\midrule
		TSDF & 0.0095  & 0.0110 & 0.0117 & 0.0104 & 0.0126 \\
		Ours w/o routing  & 0.0055 & 0.0057 & 0.0047 & 0.0055 & 0.0071 \\
		Ours & \textbf{0.0051}  & \textbf{0.0051}  & \textbf{0.0043} & \textbf{0.0052} & \textbf{0.0067} \\
		\bottomrule 
	\end{tabular}
	\vspace{-6pt}
	\caption{\textbf{Quantitative evaluation (MAD [mm]) of our method on 3D Scene Data~\cite{Zhou-et-al-SIGGRAPH-2013}}. Our method is consistently better than standard TSDF fusion on 3D Scene Data.
	These experiment also shows the benefit of our routing network when applied to real-world data.}
	\label{tab:scene3d-comparison}
	\vspace{-16pt}
\end{table}

We further show a qualitative comparison to standard TSDF as well as PSDF fusion~\cite{Dong-et-al-ECCV-2018} on the Burghers of Calais scene in Figure~\ref{fig:burghers}.
The results illustrate that our method better reconstructs fine geometric details (hands, fingers and face) and produces smoother surfaces than standard TSDF fusion and PSDF fusion~\cite{Dong-et-al-ECCV-2018}.
For more qualitative examples on this dataset, we refer to the supplementary material.

\begin{figure*}[tp]
	\vspace{-0.1cm}
	\small
	\centering
	\setlength{\tabcolsep}{0.4mm}
	\newcommand{\sz}{0.33}
	\begin{tabular}{ccc}
		\includegraphics[width=\sz\linewidth,trim={0 15 0 67},clip]{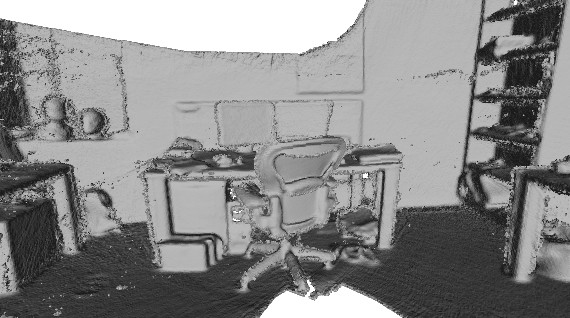} &     
		\includegraphics[width=\sz\linewidth,trim={0 15 0 67},clip]{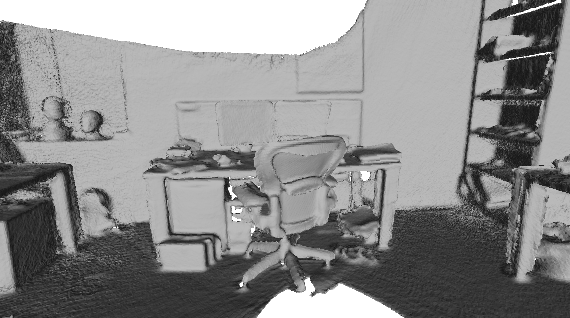} &
		\includegraphics[width=\sz\linewidth,trim={0 15 0 67},clip]{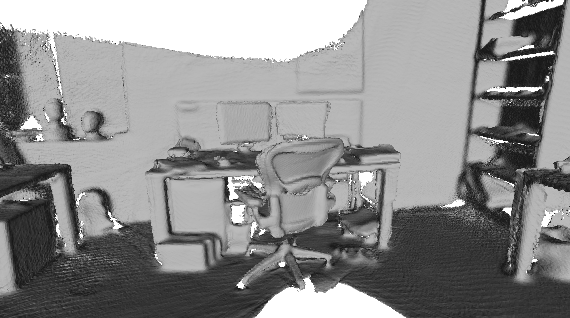} \\
		\sfc{Standard TSDF~\cite{Curless-Levoy-SIGGRAPH-1996}} & \sfc{Ours w/o Routing} & \sfc{Ours} \\[-7pt]
	\end{tabular}
	\caption{\textbf{Qualitative comparison on the heads scene of RGB-D Dataset 7-Scenes~\cite{7scene}.} Our method significantly reduces noise artifacts and thickening effects - especially on the thin geometry of the chair's leg.}
	\label{fig:microsoft-top}
	\vspace{-0.1cm}
\end{figure*}

\begin{figure}[tbp]
	\vspace{-8pt}
	\small
	\centering
	\setlength{\tabcolsep}{0.40mm}
	\newcommand{\sz}{0.31}
	\begin{tabular}{ccc}
		\includegraphics[width=\sz\columnwidth]{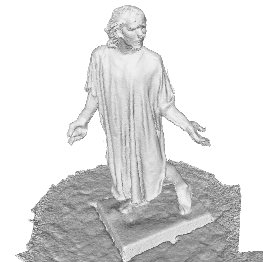} &
		\includegraphics[width=\sz\columnwidth]{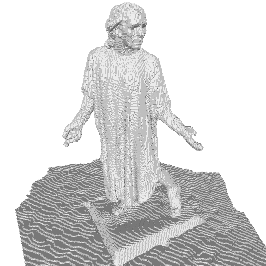} &     
		\includegraphics[width=\sz\columnwidth]{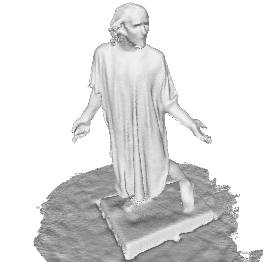}\\[0mm]
		\includegraphics[width=\sz\columnwidth,trim={0 0 0 35},clip]{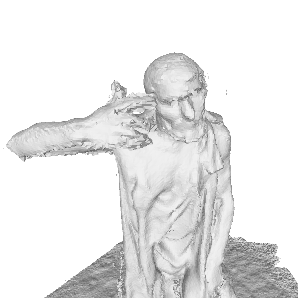} &
		\includegraphics[width=\sz\columnwidth,trim={0 0 0 35},clip]{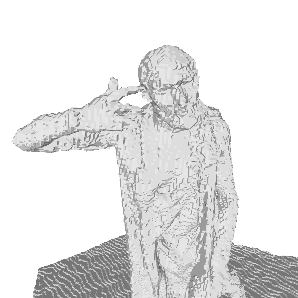} &     
		\includegraphics[width=\sz\columnwidth,trim={0 0 0 35},clip]{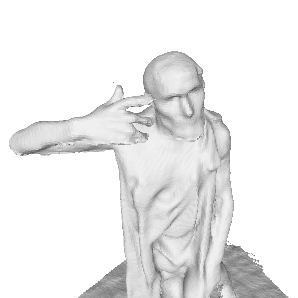}\\
		
		\sfc{TSDF \cite{Curless-Levoy-SIGGRAPH-1996}} & \sfc{PSDF \cite{Dong-et-al-ECCV-2018}} & \sfc{Ours} \\[-8pt]
	\end{tabular}	
	\caption{\textbf{Qualitative comparison on the Burghers of Calais scene~\cite{Zhou-et-al-SIGGRAPH-2013}.}
		Our method reconstructs hands and face geometry with much higher degree of detail than standard TSDF fusion and PSDF fusion.}
	\label{fig:burghers}
	\vspace{-16pt}
\end{figure}

\boldparagraph{Street Sign Dataset~\cite{Ummenhofer-Brox-ICCV-2013}.}
To evaluate the performance of our method on thin structures, we also evaluate on the street sign dataset, again without fine-tuning the network. 
This dataset consists of 50 RGB frames and we use the COLMAP SfM pipeline~\cite{schoenberger2016sfm,schoenberger2016mvs} to compute camera poses and depth maps.
Qualitative results on this scene for different state-of-the-art methods are shown in Figure~\ref{fig:street_sign_results}.
Our method clearly outperforms TV-Flux~\cite{Zach-3DPVT-2008} and standard TSDF, while producing comparable results with ray potentials~\cite{Savinov-et-al-CVPR-2016}.
The results also make the benefit of our routing network apparent.
With routing, our method reconstructs with better completeness and less noise artifacts than without. 
Note that both TV-Flux and ray potentials involve an offline optimization with a smoothness prior to reduce noise and complete missing data.
This prevents real-time application for these approaches, since ray potentials on this small scene runs for many hours on a cluster.


\boldparagraph{RGB-D Dataset 7-Scenes~\cite{7scene}.}
For qualitatively evaluating our method on Kinect data, we fuse the 7-Scenes~\cite{7scene} RGB-D dataset.
For each scene, we have chosen the first trajectory and fused it using our pipeline as well as standard TSDF fusion. 
In Figure~\ref{fig:microsoft-top}, we show that our method significantly reduces noise and mitigates the surface thickening effect compared to standard TSDF fusion. 
Notably, the chair leg and table edges are reconstructed with higher fidelity than it is done by standard TSDF fusion.
Moreover, our method shows strong performance in denoising and removing outliers from the scene.




\section{Conclusion}
%
We presented a novel real-time capable depth map fusion method tackling the common limitations of standard TSDF fusion~\cite{Curless-Levoy-SIGGRAPH-1996}.
Due to learned non-linear TSDF updates -- rather than hand-crafted linear updates -- our method mitigates inconsistent reconstruction results that occur at object edges and thin structures.
The proposed split of our network architecture into a 2D depth routing network and a 3D depth fusion network allows to effectively handle noise and outliers at different processing stages.
Moreover, sensor-specific noise distributions can be learned from small amounts of training data.
%
%
Our approach outperforms competing methods on both synthetic and real data experiments.
Due to its low computational requirements and compact architecture, our method has the potential to replace standard TSDF fusion in a variety of tasks and applications.

\noindent
\begin{minipage}{\columnwidth}
	\vspace{2mm}
	\footnotesize
	\noindent
	\textbf{Acknowledgments.}~
	Special thanks go to Akihito Seki from Toshiba Japan for insightful discussions and comments that greatly improved the paper.
This research was partially supported by Toshiba and the Intelligence Advanced Research Projects Activity (IARPA) via Department of Interior/ Interior Business Center (DOI/IBC) contract number D17PC00280.
The U.S. Government is authorized to reproduce and distribute reprints for Governmental purposes notwithstanding any copyright annotation thereon. 
Disclaimer: The views and conclusions contained herein are those of the authors and should not be interpreted as necessarily representing the official policies or endorsements, either expressed or implied, of IARPA, DOI/IBC, or the U.S. Government.
\end{minipage}

{\small
\bibliographystyle{ieee_fullname}
\bibliography{bibliography}
}



\section*{Supplementary material}

# Supplementary Material for
# RoutedFusion: Learning Real-time Depth Map Fusion

## Abstract

*In this supplementary document, we provide further visualizations and qualitative results of our learned depth map fusion approach in comparison to multiple baselines. Furthermore, we discuss the choice of different hyperparameters.*

## 1. Hyperparameters

**Number of Samples S** First, we discuss our choice for the number of samples $S$. With figure 1 and 2, we show that sampling 9 values inside the local window centered around the surface leads to the best reconstruction performance. The number of samples $S$ is closely related to the truncation distance in standard TSDF fusion [2]. Since the spacing between samples in the window is fixed to the scene's resolution, the size of the local window is dependent on $S$. Therefore, an increase in $S$ leads to an increase of the local window size. By increasing $S$ and the window size, we feed more information along the ray to the depth fusion network and we can account for larger noise levels. However, if we increase the number of samples beyond 9, the performance decreases again, which is experimentally shown in figure 1 and 2. Having empirically evaluated the influence of $S$ on our depth map fusion pipeline, we decided to keep the number of samples $S = 9$ constant across all experiments.

**Outlier Post-filtering** In order to reduce the amount of outliers in the scene, we have chosen to introduce outlier post-filtering according to the accumulated update weights during TSDF integration. Therefore, after every 100 frames integrated, we re-initialize all voxels, where the accumulated weights are smaller than 3.

## 2. Qualitative Results

In figures 3 and 4, we show more examples of reconstructions of ShapeNet [1] objects using DeepSDF [4], OccupancyNetworks [3], standard TSDF fusion [2] and our proposed method. We can clearly show that our method is superior to all other approaches shown in reconstructing these objects from noisy depth measurements.

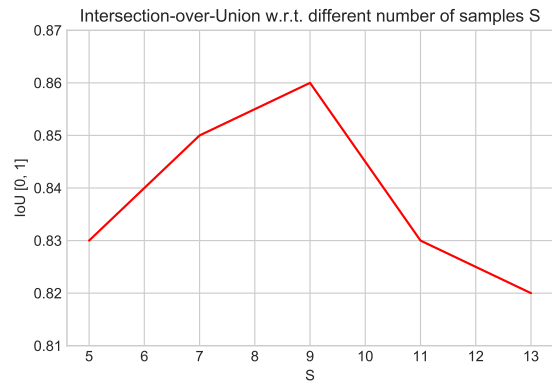

Figure 1. **Intersection over Union on Modelnet [5] test data for different numbers of samples** $S$. When sampling 9 SDF values inside the local window, our pipeline shows the best performance in reconstructing models from noisy depth measurements.

In figures 5 and 6, we qualitatively compare our method to standard TSDF fusion in reconstructing real-world scenes from the 3D scene dataset [6]. Our method significantly reduces the noise artifacts in the result, mitigates the surface thickening effect and generates very clean edges and corners.

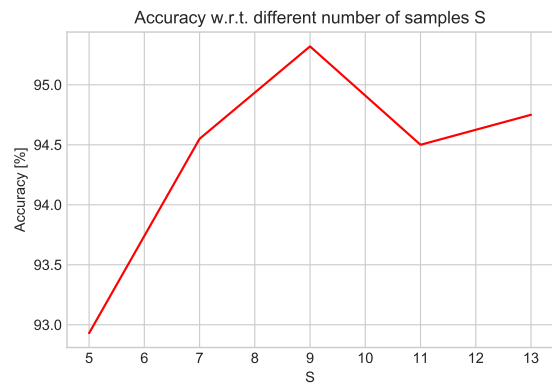

Figure 2. **Accuracy on Modelnet [5] test data for different numbers of samples** $S$. As it is the case for intersection-over-union, the accuracy peaks at 9 samples inside the window.

1

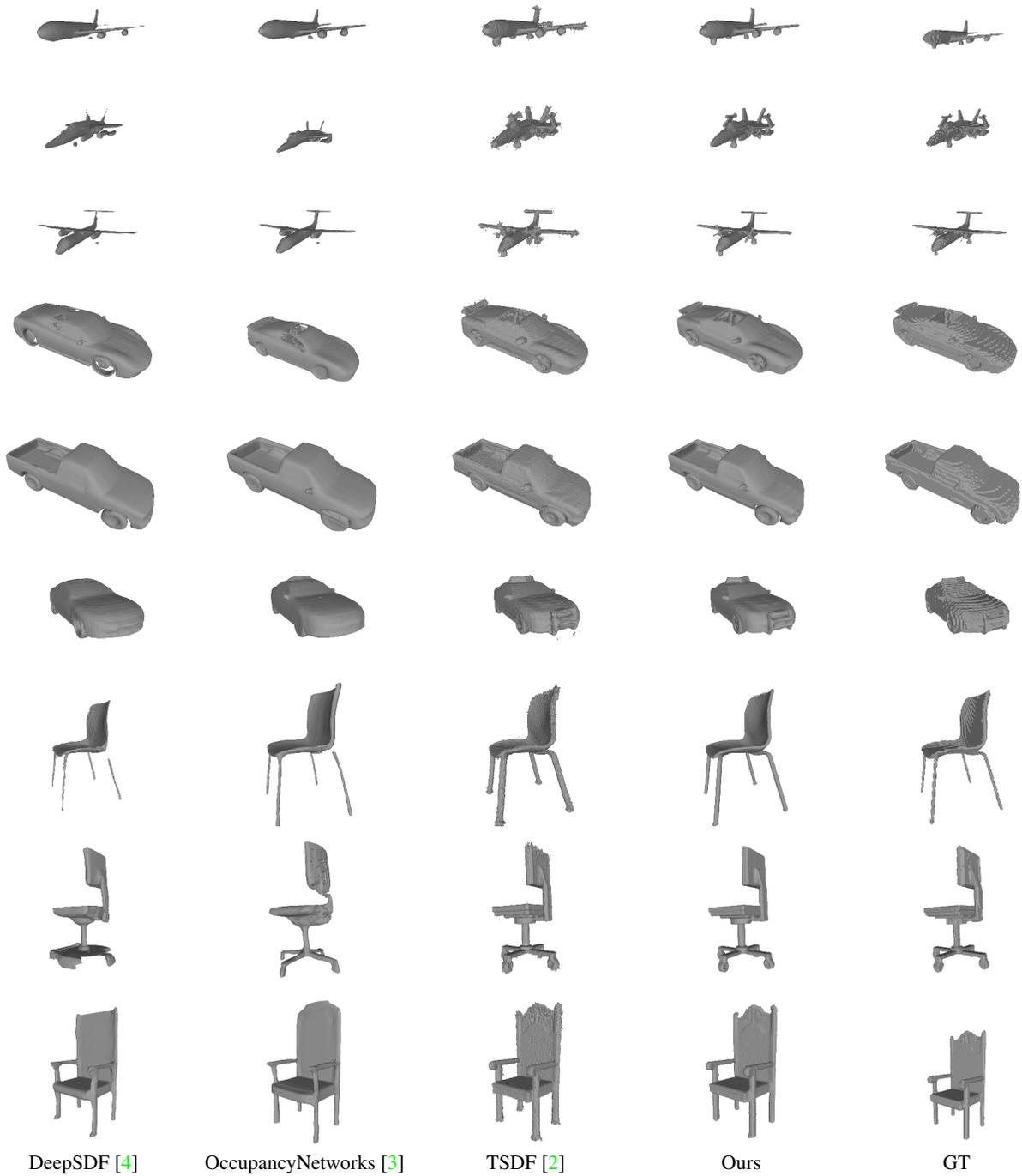

Figure 3. **More qualitative results on ShapeNet test data** They illustrate the significant performance difference in reconstructing fine geometries and clean edges between our proposed method and standard TSDF as well as recent learning-based approaches.

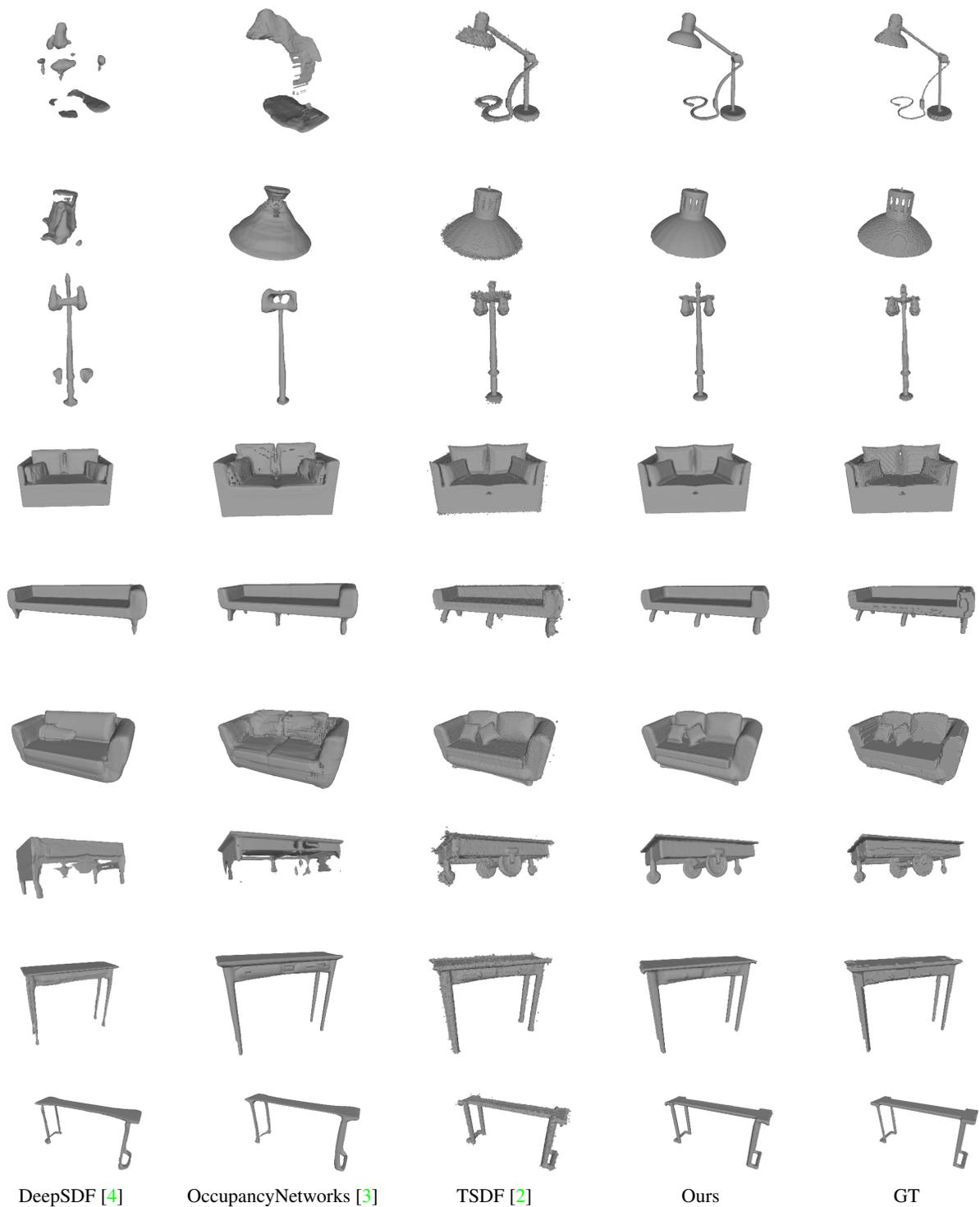

Figure 4. **More qualitative results on ShapeNet test data** They illustrate the significant performance difference in reconstructing fine geometries and clean edges between our proposed method and standard TSDF as well as recent learning-based approaches

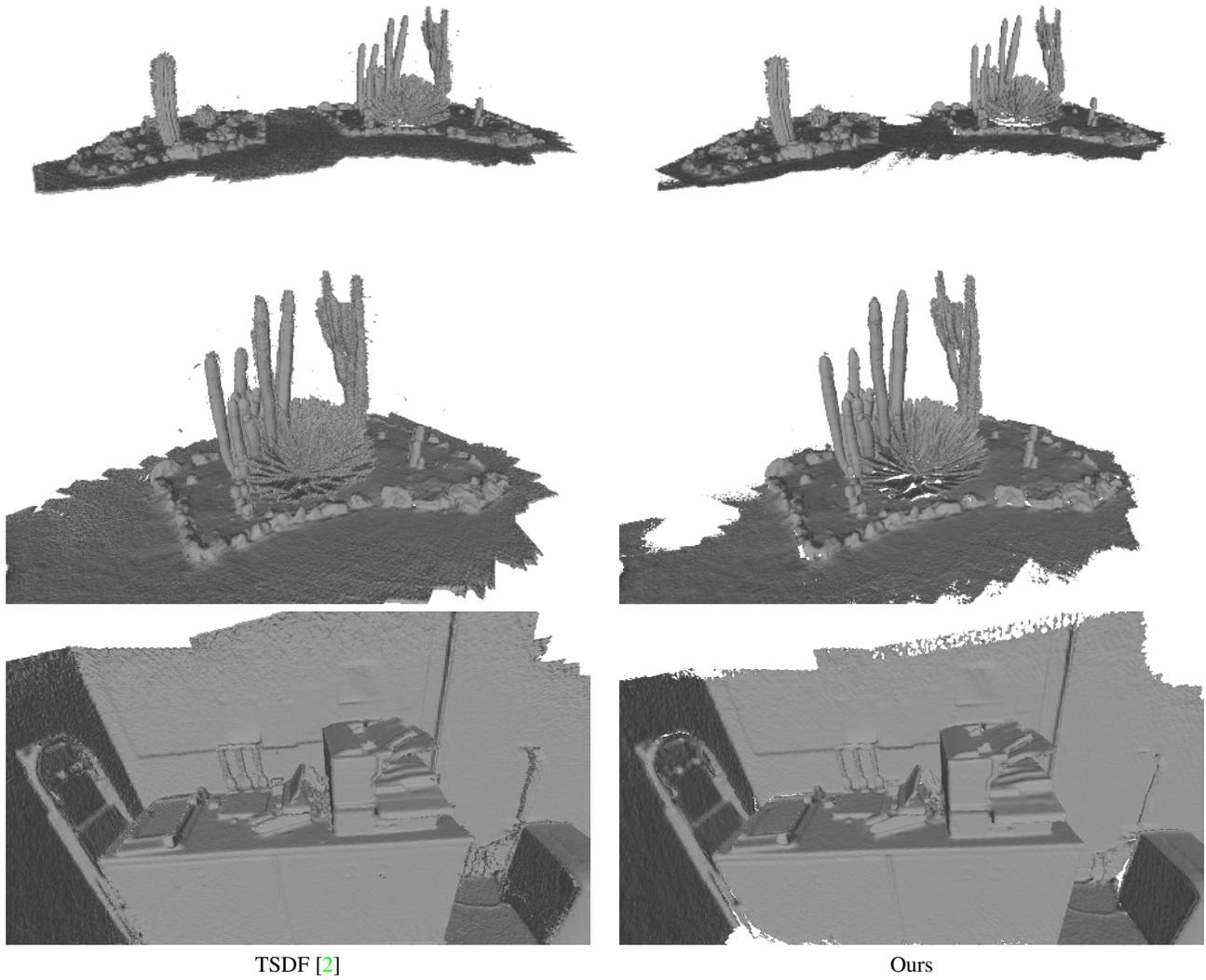

TSDF [2]           Ours

Figure 5. **More qualitative results of standard TSDF and our method on scene 3D data.** They illustrate the significant performance difference in reconstructing fine geometries and clean edges.

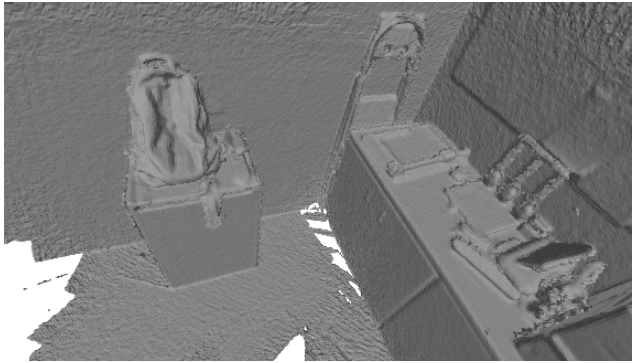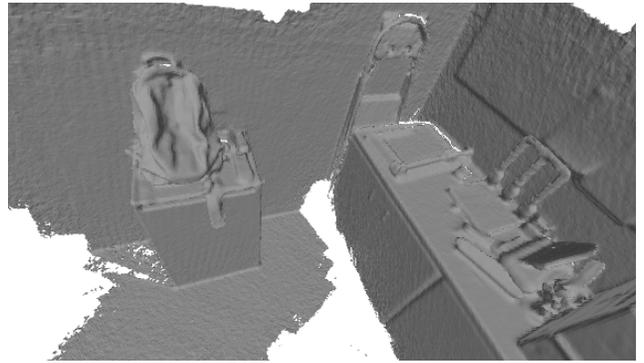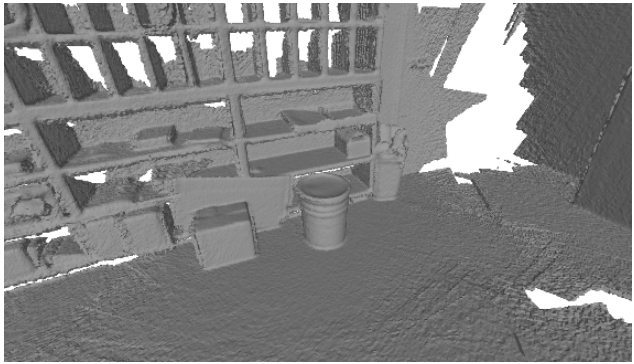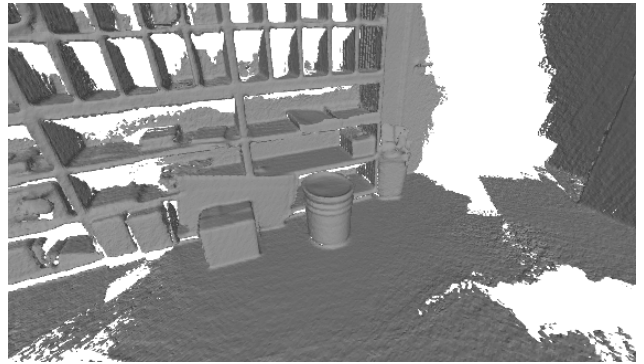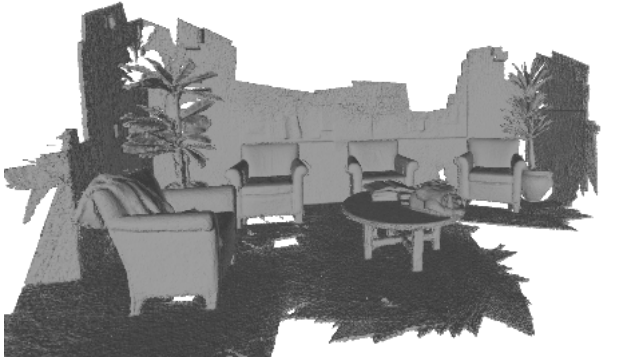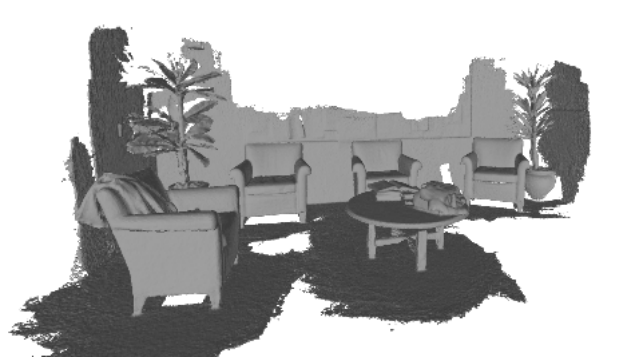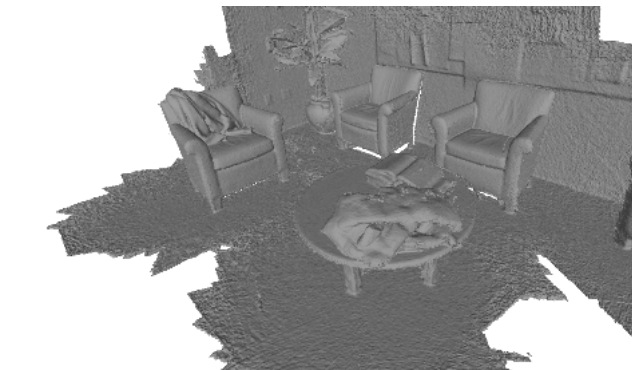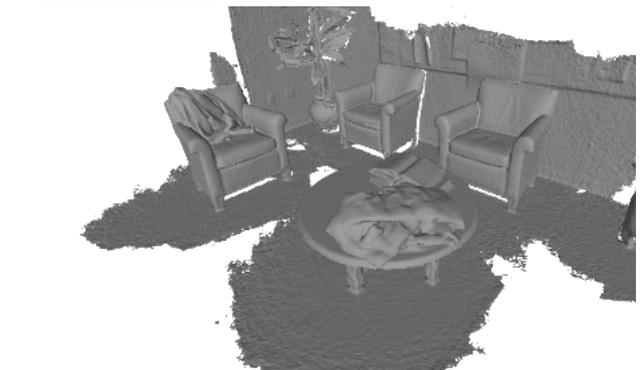

TSDF [2]          Ours

Figure 6. **More qualitative results of standard TSDF and our method on scene 3D data.** They illustrate the significant performance difference in reconstructing fine geometries and clean edges.



\end{document}